%% file: manuscript.tex
\documentclass[10pt,twocolumn,letterpaper]{article}
\pdfoutput=1
\usepackage{iccv}
\usepackage{times}
\usepackage{epsfig}
\usepackage{graphicx}
\usepackage{amsmath}
\usepackage{amssymb}

\usepackage[table]{xcolor}
\usepackage{soul}
\usepackage{booktabs}
\usepackage[normalem]{ulem}

\usepackage{algorithm,algpseudocode}
\usepackage{bbm}

\usepackage[pagebackref=true,breaklinks=true,letterpaper=true,colorlinks,bookmarks=false]{hyperref}

\iccvfinalcopy 

\begin{document}

\title{Leveraging Self-Supervision for Cross-Domain Crowd Counting}

\author{
	Weizhe Liu\textsuperscript{}
	\quad
    Nikita Durasov\textsuperscript{}
	\quad
	Pascal Fua\textsuperscript{}\\
	\textsuperscript{}Computer Vision Laboratory, \'{E}cole Polytechnique F\'{e}d\'{e}rale de Lausanne (EPFL)\\
	{\tt\small \{weizhe.liu, nikita.durasov, pascal.fua\}@epfl.ch}
}

\maketitle

\input{tex/defs}

\input{tex/abstract}
\input{tex/intro}
\input{tex/related}

\input{tex/approach}
\input{tex/experiments}

\input{tex/conclusion}

\newpage

{\small
\bibliographystyle{ieee_fullname}
\bibliography{string,vision,learning}
}

\end{document}

%% file: tex/defs.tex

\newif\ifdraft
\draftfalse

\definecolor{orange}{rgb}{1,0.5,0}
\definecolor{violet}{RGB}{70,0,170}
\definecolor{magenta}{RGB}{170,0,170}
\definecolor{dgreen}{RGB}{0,150,0}

\ifdraft
 \newcommand{\PF}[1]{{\color{red}{\bf PF: #1}}}
 \newcommand{\pf}[1]{{\color{red} #1}}
 \newcommand{\WL}[1]{{\color{blue}{\bf WL: #1}}}
 \newcommand{\wl}[1]{{\color{blue} #1}}
  \newcommand{\ND}[1]{{\color{dgreen}{\bf ND: #1}}}
 \newcommand{\nd}[1]{{\color{dgreen} #1}}
\else
 \newcommand{\PF}[1]{}
 \newcommand{\pf}[1]{#1}
 \newcommand{\WL}[1]{}
 \newcommand{\wl}[1]{#1}
 \newcommand{\ND}[1]{}
  \newcommand{\nd}[1]{#1}
\fi

\newcommand{\comment}[1]{}
\newcommand{\parag}[1]{\vspace{-3mm}\paragraph{#1}}
\newcommand{\sparag}[1]{\subparagraph{#1}}
\renewcommand{\floatpagefraction}{.99}

\newcommand{\bA}{\mathbf{A}}
\newcommand{\bC}{\mathbf{C}}
\newcommand{\bD}{\mathbf{D}}
\newcommand{\bI}{\mathbf{I}}
\newcommand{\bH}{\mathbf{H}}
\newcommand{\bP}{\mathbf{P}}
\newcommand{\bR}{\mathbf{R}}
\newcommand{\bZ}{\mathbf{Z}}

\newcommand{\real}{\mathbb{R}}

\newcommand{\bc}{\mathbf{c}}
\newcommand{\f}{\mathbf{f}}
\newcommand{\m}{\mathbf{m}}
\newcommand{\s}{\mathbf{s}}
\newcommand{\bu}{\mathbf{u}}
\newcommand{\x}{\mathbf{x}}
\newcommand{\y}{\mathbf{y}}
\newcommand{\z}{\mathbf{z}}
\newcommand{\w}{\mathbf{w}}

\newcommand{\radius}{\mathbf{r}}

\newcommand{\cF}{\mathcal F}
\newcommand{\fd}{\mathcal{F}_{d}}
\newcommand{\fz}{\mathcal{F}_{z}}

\newcommand{\OURS}[0]{\textbf{OURS}}
\newcommand{\FGSMU}[1]{\textbf{FGSM-U(#1)}}
\newcommand{\FGSMT}[1]{\textbf{FGSM-T(#1)}}
\newcommand{\FGSMUE}[1]{\textbf{FGSM-UE(#1)}}
\newcommand{\FGSMTE}[1]{\textbf{FGSM-TE(#1)}}

\newcommand{\colvecTwo}[2]{\ensuremath{
		\begin{bmatrix}{#1}	\\	{#2}	\end{bmatrix}
}}
\newcommand{\colvec}[3]{\ensuremath{
		\begin{bmatrix}{#1}	\\	{#2}	\\	{#3} \end{bmatrix}
}}
\newcommand{\colvecFour}[4]{\ensuremath{
		\begin{bmatrix}{#1}	\\	{#2}	\\	{#3} \\	{#4}	\end{bmatrix}
}}

\newcommand{\rowvecTwo}[2]{\ensuremath{
		\begin{bmatrix}{#1}	&	{#2}	\end{bmatrix}
}}
\newcommand{\rowvec}[3]{\ensuremath{
		\begin{bmatrix}{#1} &	{#2}	&	{#3} \end{bmatrix}
}}
\newcommand{\rowvecFour}[4]{\ensuremath{
		\begin{bmatrix}{#1}	&	{#2}	&	{#3} &	{#4}	\end{bmatrix}
}}

\newcommand{\tr}{^\intercal}

%% file: tex/abstract.tex

\begin{abstract}

    State-of-the-art methods for counting people in crowded scenes rely on deep networks to estimate crowd density. While effective, these data-driven approaches rely on large amount of data annotation to achieve good performance, which stops these models from being deployed in emergencies during which data annotation is either too costly or cannot be obtained fast enough. 
    
    One popular solution is to use synthetic data for training. Unfortunately, due to domain shift, the resulting models generalize poorly on real imagery.  We remedy this shortcoming by training with both synthetic images, along with their associated labels, and unlabeled real images. To this end, we force our network to learn perspective-aware features by training it to recognize upside-down real images from regular ones and  incorporate into it the ability to predict its own uncertainty so that it can generate useful pseudo labels for fine-tuning purposes. This yields an algorithm that consistently outperforms state-of-the-art cross-domain crowd counting ones without any extra computation at inference time.

\end{abstract}

%% file: tex/intro.tex

\section{Introduction}

\input{fig/intro}

Crowd counting is important for applications such as video surveillance and traffic control. For example during the current COVID-19 pandemic, it has a role to play in monitoring social distancing and slowing down the spread of the disease. Most state-of-the-art approaches rely on regressors to estimate the local crowd density in individual images, which they then proceed to integrate over portions of the images to produce people counts. The regressors typically use Random Forests~\cite{Lempitsky10}, Gaussian Processes~\cite{Chan09}, or more recently Deep Networks~\cite{Zhang15c,Zhang16c,Onoro16,Sam17,Xiong17,Sindagi17,Shen18,Liu18b,Li18f,Sam18,Shi18,Zhang20,Liu18c,Liu19g,Sam20,Idrees18,Ranjan18,Cao18}, with most state-of-the-art approaches now relying on the latter. 

Unfortunately, training such deep networks in a traditional supervised manner requires much ground-truth annotation. This is expensive and time-consuming and has slowed down the deployment of data-driven approaches. One way around this difficulty is to use synthetic data for training purposes. However there is usually too much domain shift---change in statistical properties---between real and synthetic images for networks trained in this manner to perform well, as shown in Fig.~\ref{fig:intro}.

In this paper, we remedy this shortcoming by training with both synthetic images, along with their associated labels, and {\it unlabeled} real images. We force our network to learn perspective-aware features on the real images and build into it the ability to use these features to predict its own uncertainty using a fast variant of the ensemble method~\cite{Durasov21} to effectively use pseudo labels for fine-tuning. We train it as follows:
\begin{enumerate}

 \item Initially we use synthetic images, unlabeled real images, and upside-down version of the latter. We train the network not only to give good results on the synthetic images but also to recognize if the real images are upside-up or upside-down. This simple approach to self-supervision forces the network to learn features that are perspective-aware on the real images.

 \item At the end of this first training phase in which we perform {\it image-wise} self supervision on the real images, our network is semi-trained and the uncertainties attached to the people densities it estimates have meaning. We exploit them to provide  {\it pixel-wise} self-supervision by treating the densities the network is confident about as pseudo labels, that we use as if they were ground-truth labels to re-train the network. We iterate this process until convergence. 

\end{enumerate}
Our contribution is therefore a novel approach to self-supervision for cross-domain crowd counting that relies on  {\it stochastic} density maps, that is, maps with uncertainties attached to them, instead of the more traditional  {\it deterministic} density maps. Furthermore, it explicitly leverages a specificity of the crowd counting problem, namely the fact that perspective distortion affects density counts.
We will show that it consistently outperforms the state-of-the-art cross-domain crowd counting methods.

%% file: fig/intro.tex

\begin{figure}[t]
\centering
\begin{tabular}{cc}
\includegraphics[width=.45\linewidth]{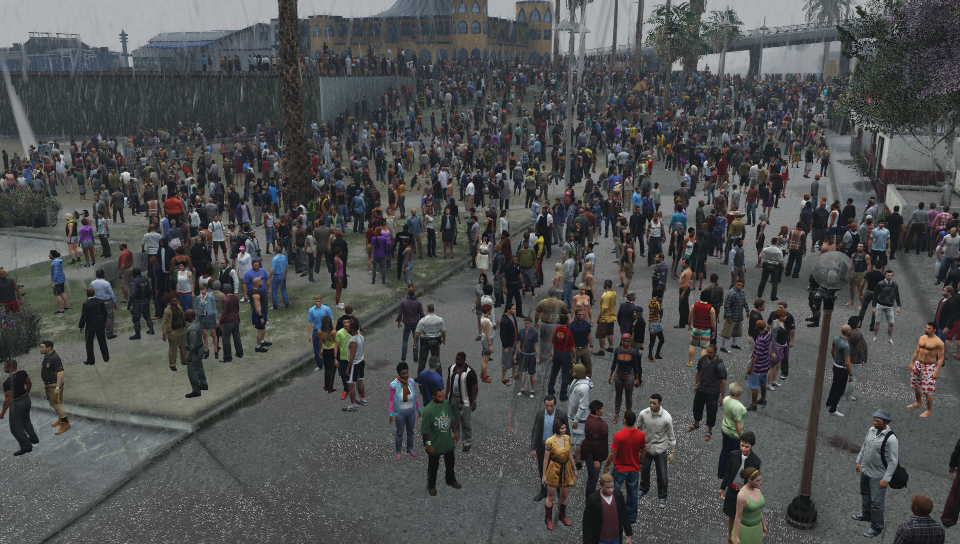}&
\includegraphics[width=.45\linewidth]{images/intro/dst.jpg}\\

\includegraphics[width=.45\linewidth]{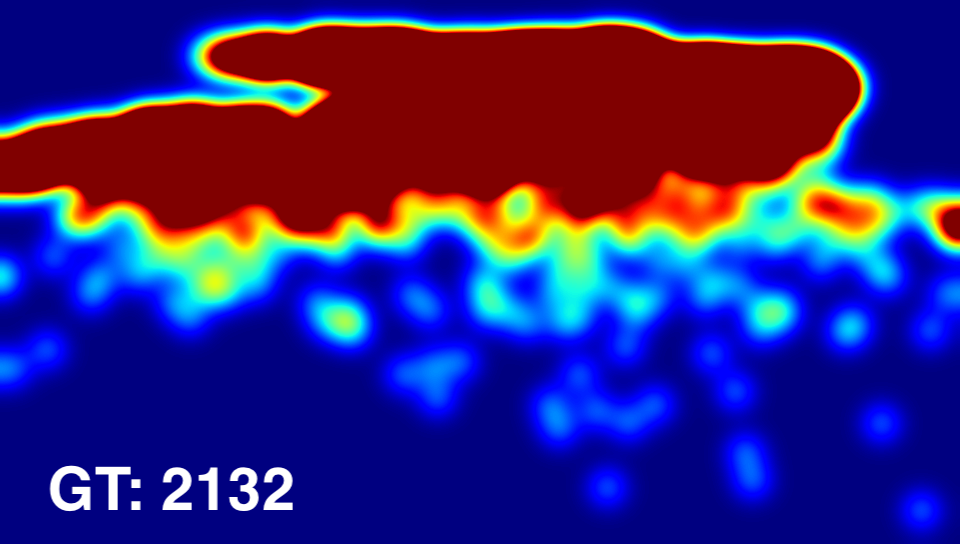}&
\includegraphics[width=.45\linewidth]{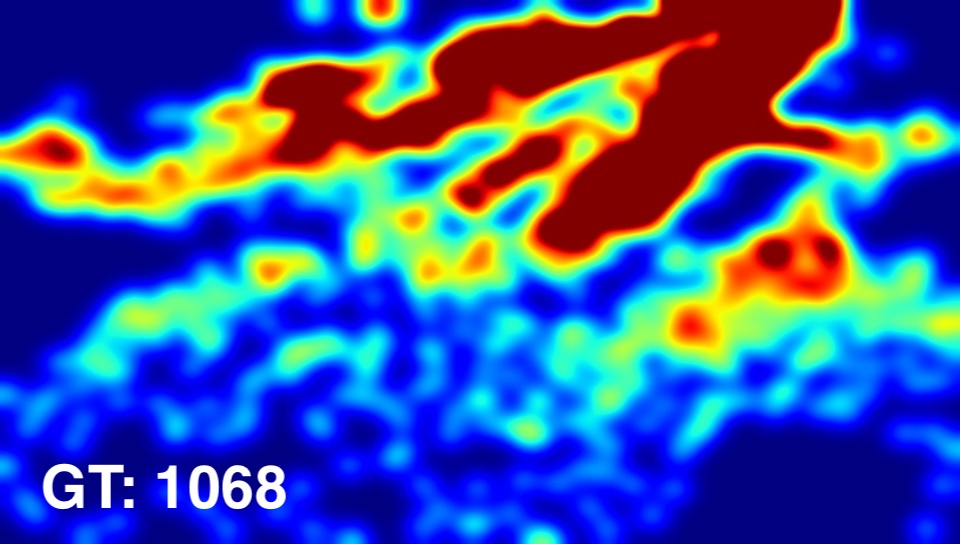}\\

\includegraphics[width=.45\linewidth]{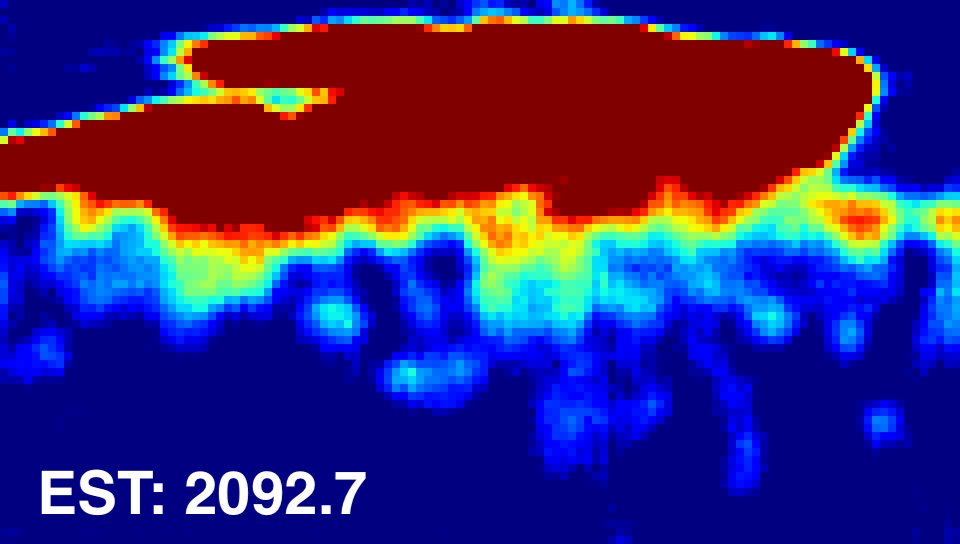}&
\includegraphics[width=.45\linewidth]{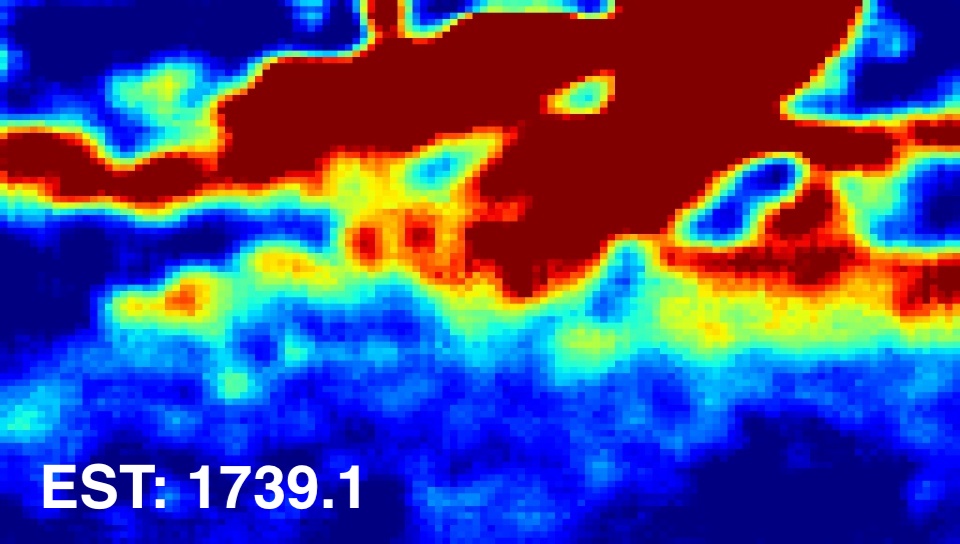}\\
\footnotesize{Synthetic}&
\footnotesize{Real}
\end{tabular}
  \caption{\small {\bf Motivation.} {\bf Top row}: Synthetic and real images unseen during training. {\bf Middle row}: Ground-truth people density maps. The total number of people obtained by integrating these maps is overlaid on the images. {\bf Bottom row}: Estimated people density maps by the network of~\cite{Wang19a} with overlaid estimated total number of people. Because the network has been trained on synthetic images, the estimated number of people in the synthetic image is very close to the correct one. This is {\it not} the case in the real one because of the large domain shift between synthetic and real images.}
  \label{fig:intro}
  \end{figure}

%% file: tex/related.tex

\section{Related Work}
\label{sec:related}

Given a single image of a crowded scene, the currently dominant approach to counting people is to train a deep network to regress a people density estimate at every image location. This density is then integrated to deliver an actual count~\cite{Liu19c,Liu19e,Shi19a,Liu19a,Jiang19a,Zhao19a,Zhang19a,Wan19a,Lian19a,Liu19d,Yan19a,Ma19a,Liu19f,Xiong19a}.  
Most methods work on counting people from individual images~\cite{Xu19a,Shi19b,Sindagi19a,Cheng19a,Wan19b,Zhang19b,Zhang19c} while others account for temporal consistency in video sequence~\cite{Xiong17, Zhang17c, Fang19a, Liu19b,Liu20a,Liu20f}. 

While effective these approaches require a large annotated dataset for training purposes, which is hard to obtain in many real-world scenarios. Unsupervised domain adaptation seek to address this difficulty. We discuss earlier approaches to it, first in a generic context and then for the specific purpose of crowd counting. 

\parag{Unsupervised Domain Adaptation.}

Unsupervised domain adaptation aims to align the source and target domain feature distributions given annotated data {\it only} in the source domain. A popular approach is to learn domain-invariant features by adversarial learning~\cite{Tzeng15, Ganin16, Hoffman16, Tzeng17, chen18d, Hong18a, Saito18b, Zhang18c, Chen19e, Zhang19e, Kurmi19, Ma19b, Cicek19, Cui20, Hu20c, Lu20b, Wu20},  which leverages one extra discriminator network to narrow the gap between two different domains. Another way to bridge the domain gap is to define a specific domain shift metric that is then minimized during training~\cite{Long15a,Long17b,Kang19, Deng19, Vu19, Pan19b, Kim19, Roy19, Lee19,Xu19c, Liang19b, Lee19b, Xu19d, Xu20, Li20e, Peng20b}. Other widely used approaches include generating realistic-looking synthetic images~\cite{Sankaranarayanan18, Hoffman18, Binkowski19, Yang20c,Yang20d}, incorporating self-training~\cite{Sener16, Chen19d, Gu20, Shin20}, transferring model weights between different domains~\cite{Rozantsev18,Rozantsev19}, and using domain-specific batch normalization~\cite{Chang19}. The method of~\cite{Sun19} introduces a self-supervised auxiliary task such as detecting image-rotation in unlabeled target domain images for cross-domain image classification and served as an inspiration to us.

\parag{Crowd Counting.}

Most of the techniques described above are intended for classification problems and very few have been demonstrated for crowd counting purposes. 

One exception is the method of~\cite{Wang19a,Gao19b,Wang20d} that trains the deep model on synthetic images and then narrows the domain gap, by using a CycleGAN~\cite{Zhu17a} extension to translate synthetic images to make them look real and then re-train the model on these translated images. A limitation of this work is that the translated images, while more realistic than the original synthetic ones, are still not truly real.

Another exception is the method of~\cite{Sindagi20a}. It uses pseudo labels generated by a network trained on synthetic images as though they were ground-truth labels. It relies on Gaussian Processes 
to estimate the variance of the pseudo labels and to minimize it. However, the uncertainty of these pseudo labels is not estimated or taken into account and the computational requirements can become very large when many synthetic images are used simultaneously.  

The method of~\cite{Han20} uses adversarial learning to align features across different domains. However, it relies on extra discriminator networks which are complicated and hard to train. ~\cite{Reddy20,Hossain19,Wang21} leverage a few target labels to bridge the domain gap, therefore require extra annotation cost.

By contrast to these approaches, ours explicitly takes uncertainty into account and leverages a specificity of the crowd counting problem, namely the fact that perspective distortion matters.

%% file: tex/approach.tex

\section{Approach}
\label{sec:approach}

We propose a fully unsupervised approach to fine-tuning a network that has been trained on annotated synthetic data, so that it can operate effectively on real data despite a potentially large domain shift. 
At the heart of our method is a network that estimates people-density  at every location while incorporating a variant of the deep ensemble approach~\cite{Durasov21} to provide uncertainties about these. The key to success is to first pre-train this network so that these uncertainties are meaningful and then to exploit them to recursively fine-tune the network.

\input{fig/model}

We have therefore developed a two-stage approach that first relies on real-images and upside-down versions of these to provide an {\it image-wise} supervisory signal. We use them to train the network not only to give good results on the synthetic images but also to recognize if the real images are upside-up or upside-down. This yields a partially-trained network that can operate on real images and return meaningful uncertainty values along with the density values. We can therefore exploit them to provide  {\it pixel-wise} supervisory signal, by treating the people density estimates the network is most confident about as pseudo labels, that are treated as ground-truth and use to re-train the network. We iterate this process until the network predictions stabilize. Fig.~\ref{fig:model} depicts our complete approach.

\subsection{Network Architecture}

Formally, let $D^{s} = \{(\x^{s}_{i},\y^{s}_{i})\}_{i=1}^{N_{s}}$ be a synthetic source-domain dataset, where $\x^{s}$ denotes a color synthetic image and $\y^{s}$ the corresponding crowd density map.  The target-domain dataset is defined as $D^{t} = \{\x^{t}_{i}\}^{N_{t}}_{i=1}$ without ground truth crowd density labels where $\x^{t}$ denotes a color real image. In most real-world scenarios, we have $N_{s} \gg N_{t}$.  Our goal is to learn a model that performs well on the  target-domain data. 

To this end, we use a state-of-the-art encoder/decoder architecture for people density estimation~\cite{Wang19a}. We chose this one because it has already been used by  cross-domain crowd counting approaches and therefore allows for a fair comparison of our approach against earlier ones.  Let ${\cal E}$ and ${\cal D}$ be the encoder and decoder networks that jointly form the people density estimation network ${\cal F}$ of~\cite{Wang19a}.  Given an input image $\x$ as input,  ${\cal E}$  returns the deep features $\f={\cal E}(\x)$ that  ${\cal D}$ takes as input to return the density map ${\cal D}(\f)$.

One way to enable self-supervision for classification purposes is to use a partially trained network to predict labels and associated probabilities, treat the most probable ones as {\it pseudo labels} that can be used for training purposes as though they were ground-truth labels~\cite{Yang20c,Yang20d}. This strategy is widely used to provide pixel-wise~\cite{Zou18} and image-wise~\cite{Zou19} self-supervision to address classification problems. If the probability measure is reliable and allows the discarding of potentially erroneous labels, repeating this procedure several times results in the network being progressively refined without any need for ground-truth labels. 

\input{fig/mask}

To implement a similar mechanism in our context, we need more than labels at the image-level. We require estimates of which individual densities in an estimated density map are likely correct and which are not. In other words, we need a  {\it stochastic} crowd density map instead of the deterministic one that existing methods produce. Among all the methods that can be used to turn our network $\cF$ into one that returns such stochastic density maps, MC-Dropout~\cite{Gal16} and Deep Ensembles~\cite{Lakshminarayanan17} have emerged as two of the most popular  ones. Both of those methods exploit the concept of \textit{ensembles}  to produce uncertainty estimates. Deep Ensembles are widely acknowledged to yield significantly more reliable uncertainty estimates~\cite{Ovadia19,Ashukha20}. However, they require training many different copies of the network, which can be very slow and  memory consuming. Instead, we rely on {\it Masksembles}, a recent approach~\cite{Durasov21} that operates on the same basic principle as MC-Dropout. However, instead of achieving randomness by dropping different subsets of weights for each observed sample, it relies on a set of pre-computed binary masks that specify the network parameters to be dropped. Fig.~\ref{fig:mask} depicts this process. 

In practice, we associate to the first convolutional layer of the decoder $\cal D$ a Masksembles layer. During training, for each sample in a batch we randomly choose one of the masks, set the corresponding weights to one or zero in the Masksembles layers, which drops the corresponding parts of the model just like standard dropout. During inference, we run the model multiple times, once per mask,
to obtain a set of predictions and, ultimately, an uncertainty estimate. This turns out to provide uncertainty estimates that are almost as reliable as those of Ensembles but without having to train multiple networks and is therefore much faster and easier to train. Formally, we write
\begin{align}
     \bar{\y} & = \frac{1}{M}\sum_{m=1}^{M} {\cal F}_{m} (\x)\;,                          \label{eq:ensemble}     \\
     \bu & = \sqrt{\sum_{m=1}^{M} ({\cal F}_m (\x) - \bar{\y})^2} \; ,                    \label{eq:std}
\end{align}
where $\x$ is the input image,  ${\cal F}_{m}$ is the modified network  ${\cal F}$ used with mask $m$.  $\bar{\y}$ and  $\bu$ are the same size as input image and we treat the individual values of $u \in \bu$ as pixel-wise uncertainties. 

\subsection{Image-Wise Self-Supervision}

\input{fig/flip}

${\cal F}_{m}$ can be trained in a supervised fashion using the synthetic training set $D^{s}$ but that does not guarantee that it will work well on real images.  Hence, we introduce the auxiliary task decoder  ${\cal D}_{aux}$ shown at the top of Fig.~\ref{fig:model} whose task is to classify an image as being oriented normally or being upside-down from the features produced by the encoder. To train the resulting two-branch network, we use synthetic images from $D^{s}$ along with real images from $D^{t}$ and flipped versions of these, such as the ones shown in Fig.~\ref{fig:flip}. For the synthetic images, the output should minimize the usual $L_2$ loss given the ground-truth density maps and, for the real images, the output should minimize a cross entropy loss for binary classification as being either upside-up or upside-down. 

Formally, we introduce the loss function
\begin{align}
 L_{st1}  & =  L_{s} + \lambda_{1} L_{a}\;,  \label{eq:st1} \\
 L_{s}     & =  \sum_{i} \|\y^{s}_{i}-{\cal D}({\cal E}(\x_{i}^{s}))\|^{2} \;,   \nonumber  \\
 L_{a}     & =  -\sum_{i}\left \langle \y_{i}^{t},log({\cal D}_{aux}({\cal E}(\x_{i}^{t}))) \right \rangle \;,\nonumber
\end{align}
which we minimize with respect to the weights of the encoder ${\cal E}$ and the two decoders ${\cal D}$ and ${\cal D}_{aux}$. $L_{s}$ is the $L_2$ distance between the predicted people density map and the ground truth one $y^{s}_{i}$ while $L_{a}$ is the cross-entropy loss for binary classification given the ground-truth upside-up or down label $ \y_{i}^{t}$ for image $\x_{i}^{t}$. We use this label only for the real images because we have ground truth annotations for the synthetic ones. As will be shown in the results section, this provides sufficient supervision for the synthetic images and also using the image-wise supervision for these brings no obvious improvement. 

Note that the $L_{s}$ and $L_{a}$ use the same encoder ${\cal E}$. To minimize $L_{a}$ and hence correctly estimate if an input image is upside-down or not, ${\cal E}$ must extract meaningful features from the real images and not only from synthetic ones.  Furthermore, these features must enable the decoder ${\cal D}$ to  handle scene perspective, that is, the fact that people densities are typically higher at the top of the image than the bottom in upside-up images. In other words, minimizing $L_{a}$ forces ${\cal E}$ to produce perspective-aware features while minimizing $L_{s}$ forces the decoder ${\cal D}$ to operate on such features to properly estimate people densities on the synthetic images. In this way, we make ${\cal E}$ produce features that are appropriate both for  synthetic and real images, hence mitigating the domain shift between the two, as will be demonstrated in the results section. 

\input{table/alg_train}

This first training stage is summarized by the first procedure of Alg.~\ref{alg:train}.

\subsection{Pixel-Wise Self-Supervision}

After the first training stage described above, our model can produce both a density map  $\bar{\y}$ and its corresponding uncertainty $\bu$. Let ${\cal F}_{m}^0$ be the corresponding network. We can now refine its weights to create increasingly better tuned networks ${\cal F}_{m}^k$ for $1 \leq k \leq K$ by iteratively minimizing 
\begin{align}
  L_{st2}   & =  \sum_{i} \|\y^{s}_{i}-{\cal F}_{m}^k(\x_{i}^{s})\|^{2}  \label{eq:st2} \\
                & \quad +  \lambda_{2} \sum_{i} \| \mathbbm{1}_{u^{k-1}_i < u_{\alpha}} (\bar{\y}^{k-1}_{i}-{\cal F}_{m}^k(\x_{i}^{t}) )\|^{2}\;, \nonumber    
\end{align}
where $\bar{\y}^{k-1}_{i},\bu^{k-1}_{i} = {\cal F}_{m}^{k-1} (\x_{i}^{t})$ and $\mathbbm{1}_{u^{k-1}_i < u_{\alpha}}$ is one for all densities for which the uncertainty is less than the top $\alpha \%$ uncertainty $u_{\alpha}$. In other words, at each iteration we use the densities produced by ${\cal F}_{m}^{k-1}$ for which the uncertainty is low enough as pseudo labels to train ${\cal F}_{m}^{k}$.

This second training stage is summarized by the second procedure of Alg.~\ref{alg:train}.

%% file: fig/model.tex

\begin{figure*}[t]
\centering
\includegraphics[width=1.0\linewidth]{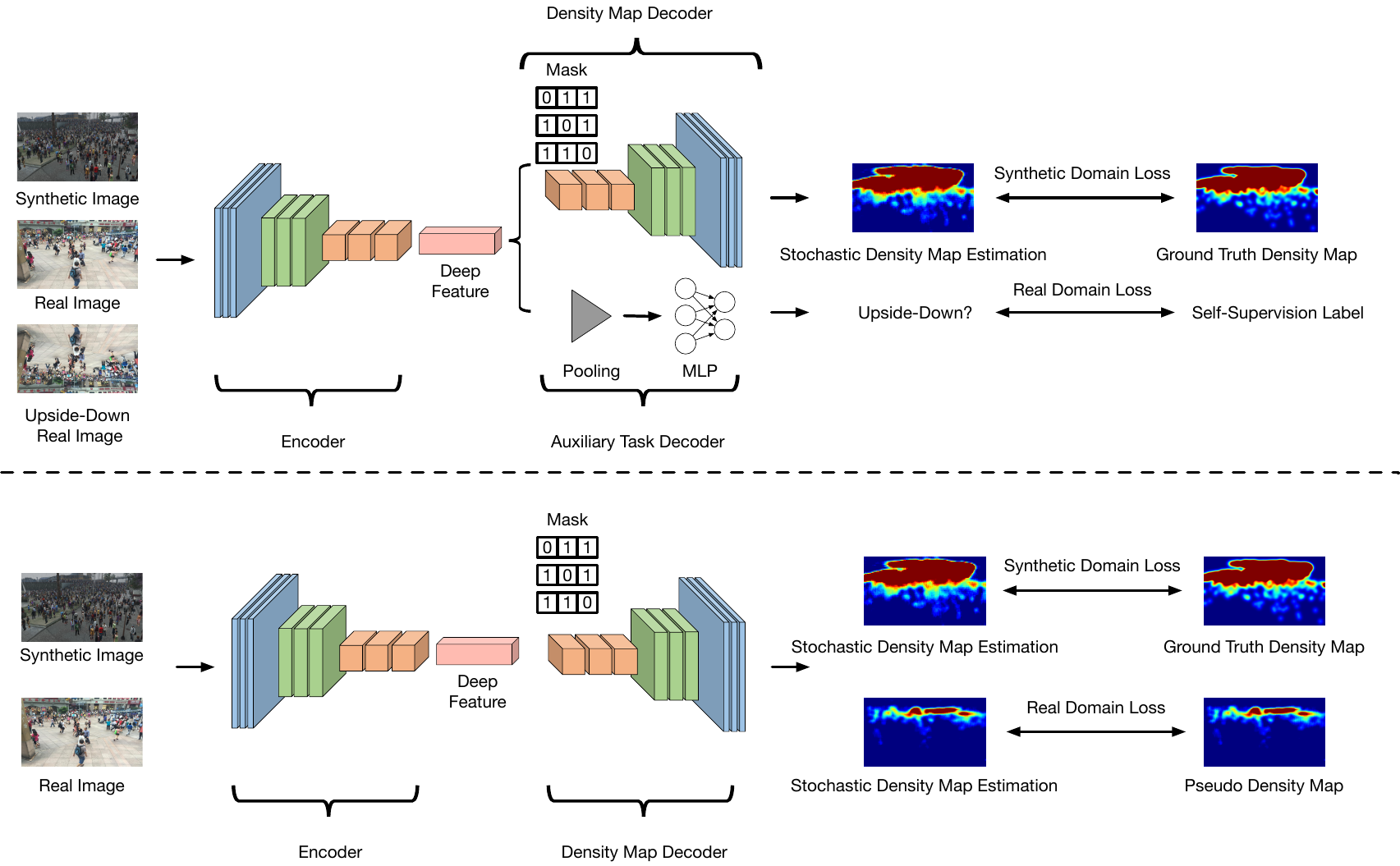}
  \caption{\small {\bf Two-stage approach.} {\bf Top:} During the first training stage, we use synthetic images, real images, and flipped versions of the latter. The network is trained to output the correct people density for the synthetic images and to classify the real images as being flipped or not. {\bf Bottom:} During the second training stage, we use synthetic and real images. We run the previously trained network on the real images and treat the least uncertain people density estimates as pseudo labels. We then fine tune the network on both kinds of images and iterate the process.  }
  \label{fig:model}
  \end{figure*}

%% file: fig/mask.tex
\begin{figure}[t]
    \centering
    \includegraphics[width=8cm]{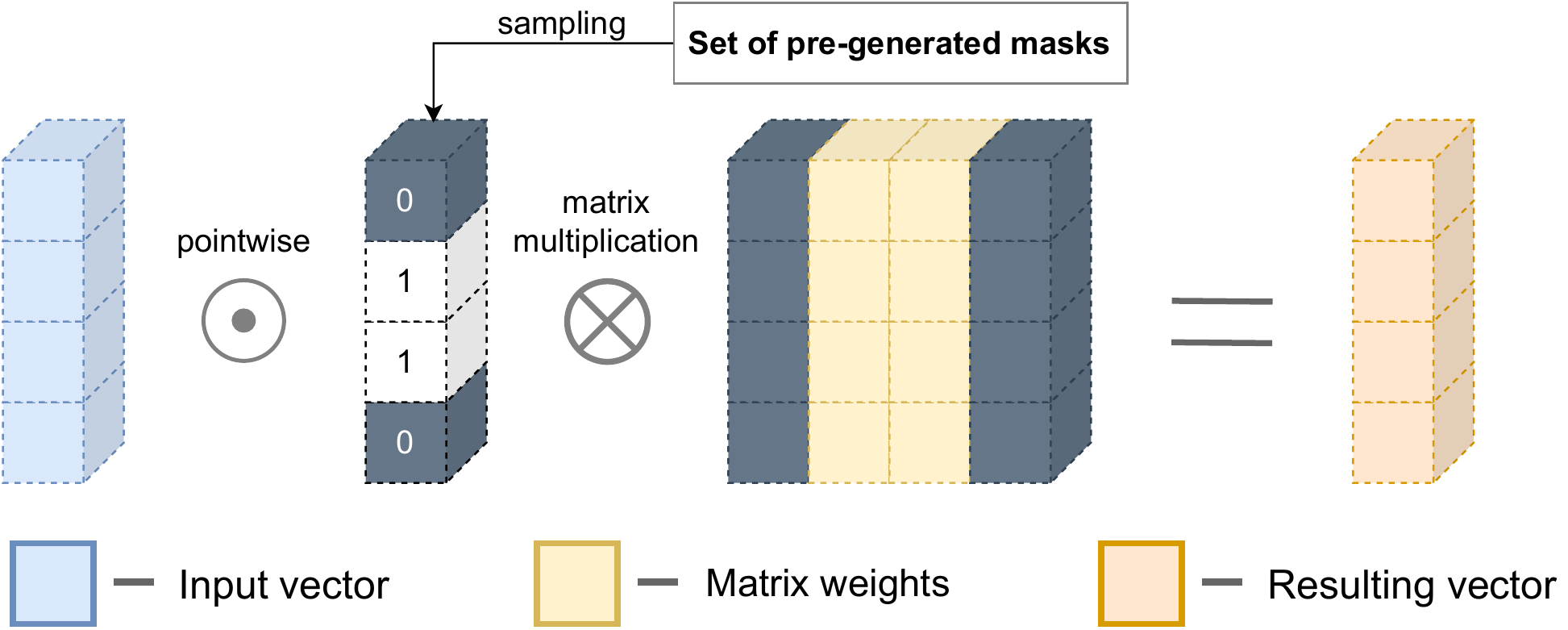}
    \caption{\small \textbf{Masksembles approach.} During training, for every input vector, a binary mask is selected from a set of pre-generated masks and is used to zero out a corresponding set of features. Performing the inference several times using different masks then yields an ensemble-like behavior.}
    \label{fig:mask}
  \end{figure}

%% file: fig/flip.tex

\begin{figure}[t]
\centering
\begin{tabular}{cc}
\includegraphics[width=.45\linewidth]{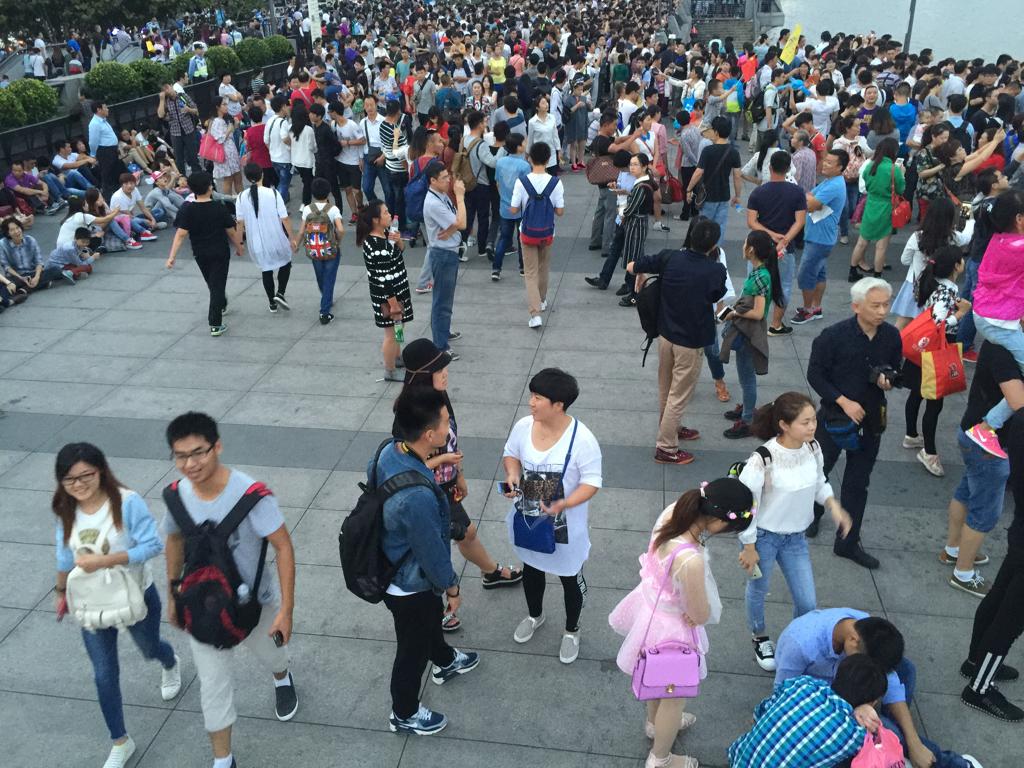}&
\includegraphics[width=.45\linewidth]{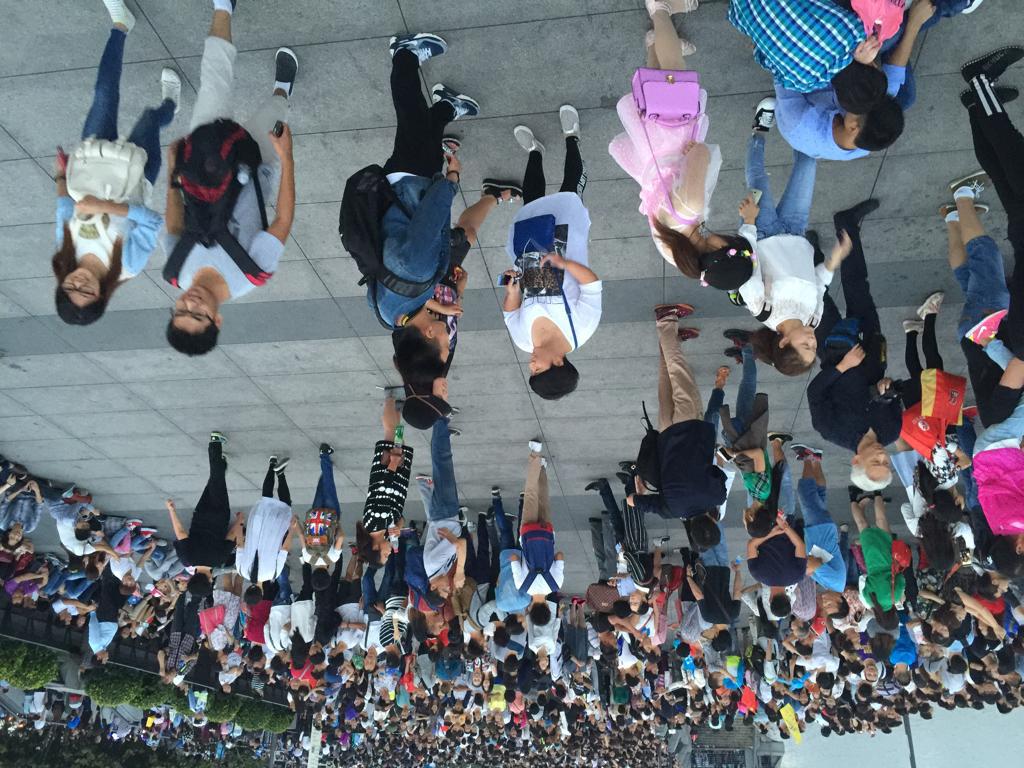}\\
\footnotesize{(a)}&
\footnotesize{(b)}
\end{tabular}
  \caption{\small {\bf Upside-up vs Upside-down.} (a) Original image. Due to perspective effects, the 2D projection of people is smaller at the top of the image and the people density appears to be larger. (b) In the upside-down image the effect is reversed. To allow the decoder to distinguish between these two cases, the encoder must produce perspective-aware features that can operate in the real images.
 }
  \label{fig:flip}
  \end{figure}

%% file: table/alg_train.tex

\begin{algorithm}[t]
    \begin{algorithmic}[0]
    \Require  Source domain data $D^{s} = \{(\x^{s}_{i},\y^{s}_{i})\}_{i=1}^{N_{s}}$ .
    \Require  Unlabeled target domain data $D^{t} = \{\x^{t}_{i}\}^{N_{t}}_{i=1}$. 
    \Statex
    \Procedure{First Stage}{ $D^{s}$ and $D^{t}$}
    \State Initialize the weights for people density estimation network ${\cal F}_{m}$ with single encoder ${\cal E}$ and two decoders ${\cal D}$ and ${\cal D}_{aux}$
    \For{$\#$ of gradient iterations}
    \State Pick one source domain image $\x^{s}_{i}$
    \State Pick one target domain image $\x^{t}_{i}$
    \State Generating one random variable $\beta \in [0,1]$
    \If{$\beta \ge 0.5$ } 
    \State Flip $\x^{t}_{i}$ upside-down
    \Else
    \State Do nothing
    \EndIf
    \State Minimize $L_{st1}$ of Eq.~\ref{eq:st1}
    \EndFor
    \EndProcedure
    \Statex
    \State Generating pseudo labels for $\x^{t}_{i} \in D^{t}$ using ${\cal F}_m$
    \Statex
    \Procedure{Second Stage}{ $D^{s}$, $D^{t}$ and pseudo labels for $\x^{t}_{i} \in D^{t}$}
    \For{$\#$ of recursive iterations}
    \For{$\#$ of gradient iterations}
    \State Pick one source domain image $\x^{s}_{i}$
    \State Pick one target domain image $\x^{t}_{i}$
    \State Minimize $L_{st2}$ of Eq.~\ref{eq:st2}
    \EndFor
    \State Update pseudo labels 
    \EndFor
    \EndProcedure
    \end{algorithmic}
    \caption{Two-Stage Training Algorithm} \label{alg:train}
\end{algorithm}

%% file: tex/experiments.tex

\newcommand{\ours}[0]{{\bf OURS}}
\newcommand{\base}[0]{{\bf BASELINE}}
\newcommand{\ouri}[0]{{\bf OURS-IMG}}
\newcommand{\ouris}[0]{{\bf OURS-IMG-SYN}}
\newcommand{\ourp}[0]{{\bf OURS-PIX}}
\newcommand{\ourd}[0]{{\bf OURS-DUP}}
\newcommand{\ourm}[0]{{\bf OURS-MIRROR}}
\newcommand{\ourn}[0]{{\bf OURS-90}}
\newcommand{\ourt}[0]{{\bf OURS-270}}
\newcommand{\supe}[0]{{\bf SUPERVISED}}

\section{Experiments}
\label{sec:experiment}

In this section, we first introduce the evaluation metrics and benchmark datasets we use in our experiments. We then provide the implementation details and compare our approach to state-of-the-art methods. Finally, we perform a  detailed ablation study.

\input{fig/vis}

\subsection{Evaluation Metrics}
\label{sec:metrics}

Previous works in crowd density estimation use the mean absolute error ($MAE$) and the root mean squared error ($RMSE$) as evaluation metrics~\cite{Wang19a,Sindagi20a}. They are defined as  
\begin{small}
\begin{equation}
    MAE = \frac{1}{N}\sum_{i=1}^{N}|z_{i}-\hat{z_{i}}| \mbox{ and } RMSE=\sqrt{\frac{1}{N}\sum_{i=1}^{N}(z_{i}-\hat{z_{i}})^{2}} \; , \nonumber
\end{equation}
\end{small}
where $N$ is the number of test images, $z_{i}$ denotes the true number of people inside the ROI of the $i$th image and $\hat{z_{i}}$ the estimated number of people. In the benchmark datasets discussed below, the ROI is the whole image except when explicitly stated otherwise. The number of people are recovered by integrating over the pixels of the predicted density maps.

\subsection{Benchmark Datasets}

\parag{GCC~\cite{Wang19a}.} 
It is the synthetic dataset we use. It consists of 15,212 images of size $1080 \times 1920$, containing 7,625,843 people annotations. It features 400 different scenes including both indoor and outdoor ones.

\parag{ShanghaiTech~\cite{Zhang16c}.} 
It is a real image dataset that comprises 1,198 annotated images with 330,165 people in them. It is divided in part A with 482 images and part B with 716. In part A, 300 images form the training set and, in part B, 400. The remainder are used for testing purpose. 

\parag{UCF-QNRF~\cite{Idrees18}.} 
It is a real image dataset that comprises 1,535 images with 1,251,642 people in them. The training set comprises 1,201 of these images. Unlike in {\bf ShanghaiTech}, there are dramatic variations both in crowd density and image resolution.

\parag{UCF\_CC\_50~\cite{Idrees13}.} 
It is a real image dataset that contains only 50 images with a people count ranging from 94 to 4,543, which makes it challenging for a deep-learning approach. For a fair comparison, we use the same 5-fold cross-validation protocol as in~\cite{Wang19a,Sindagi20a}: We partition the images into 5 10-image groups. In turn, we then pick four groups for training and the remaining one for testing. This gives us 5 sets of results and we report their average.

\parag{WorldExpo'10~\cite{Zhang15c}.} 
It is a real image dataset that comprises 1,132 annotated video sequences collected from 103 different scenes. There are 3,980 annotated frames, with 3,380 of them used for training purposes.
Each scene contains a Region Of Interest (ROI)  in which people are counted. As in previous work~\cite{Wang19a}, we report the \textit{MAE} for each scene along with the average over all scenes.

\input{table/eva}

\subsection{Implementation Details}

For a fair comparison with previous work~\cite{Wang19a,Sindagi20a}, we use SFCN~\cite{Wang19a} as the crowd density regressor and Adam~\cite{Kingma15} for parameter update with a learning rate of $1e-6$. After a grid search on {\it one single} dataset as discussed below, we set $\lambda_{1}$ in Eq.~\ref{eq:st1}, $\lambda_{2}$, and $K$ in Eq.~\ref{eq:st2} to $10^{-4}$, $1.0$ and $2$ respectively for all our experiments. 

To estimate uncertainty, we generate $3$ stochastic density map for each image and take the standard deviation to be our uncertainty measure. We set the threshold value $\alpha$ of Eq.~\ref{eq:st2} to $10$, which means that $10\%$ most uncertain pseudo labels are discarded and that we keep the other $90\%$ as pseudo labels for model training. This large percentage is appropriate because there are large areas of the real images that do not contain anyone and for which the pseudo labels are very dependable. We will show below that removing only 10\% of the labels suffices to substantially boost performance over keeping all pseudo labels.

Recall that we drop the auxiliary network ${\cal D}_{aux}$ in the second training stage.  In the final evaluation phase, we generate only one density map for each image instead of averaging multiple estimates, we will show that the performance is similar for both cases in supplementary material. Hence our model does not require any extra computation at inference time. Fig.~\ref{fig:vis} depicts qualitative results on ShanghaiTech Part B dataset and we provide additional ones in the supplementary material along with more details about the model. 

\subsection{Comparing against Recent Techniques}

In Tab.~\ref{tab:eva}, we compare our results to those of state-of-the-art domain adaptation approaches for each one of the public benchmark datasets, as currently reported in the literature. In each case, we reprint the results as given in these papers and add those of \ours{}, that is, of our method. We consistently and clearly outperform all other methods on all the datasets. And, since we use the same SFCN network architecture as the methods of~\cite{Wang19a,Sindagi20a}, the performance boost is directly attributable to our approach of domain adaptation. 

In~\cite{Wang19a}, the authors report {\it fully supervised} MAE results on  Shanghaitech Part B and UCF-QNRF of 9.4 and 124.7, respectively, to be compared to our own {\it unsupervised} values of 11.4 and 198.3. In other words, our unsupervised approach performs almost as well as a supervised one on Shanghaitech Part B  while there still remains a gap on UCF-QNRF. This is because the crowds in both the synthetic source domain and in Shanghaitech Part B are still mostly sparse enough for bodies to be visible. By contrast, in UCF-QNRF, the crowds are denser. Hence, it often happens that only heads are visible, thus creating a larger domain gap between source and target images that could be bridged in future work either by using a synthetic dataset that itself features denser crowds or, more ambitiously, by using a detection pipeline that focuses more on heads and would naturally reduce the domain gap. 

\subsection{Ablation Study}
\label{sec:ablation}

We perform an ablation study on UCF-QNRF dataset to confirm the role of  the self-supervision loss terms, the setting of hyper-parameters, the impact of stochastic density map, the choice of auxiliary task and to compare against other uncertainty estimation techniques. 

\parag{Self-Supervision.} 

We compare our complete model against several variants. \base{} uses the SFCN crowd density estimator trained on the synthetic data and without any domain adaptation. \ouri{} involves the first image-wise training stage but not the second. \ouris{} also involves only the first image-wise training stage but both real and {\it synthetic} images can be flipped upside down, whereas in \ouri{} only the real ones are. Conversely,  \ourp{} skips the first image-wise training and involves only the second pixel-wise training stage.  \ourd{} is similar to our complete approach except for the fact that it uses both pixel-wise and image-wise supervision during the second training stage whereas \ours{} only uses pixel-wise supervision by that point. 

\input{table/ab_sup}

As shown in Tab.~\ref{tab:ab_supervision}, both \ouri{} and \ourp{} outperform \base{} which shows that both training stages matter. However, \ours{} does even better, which confirms that properly pre-training the network before using pixel-wise supervision matters. Since \ouris{} and \ourd{} achieve similar performance as \ouri{} and \ours{} respectively, we drop image-wise self-supervision for synthetic image and in the second stage for simplicity.

\input{table/ab_parameter}

\parag{Hyper-Parameter Selection.} 

We tested different values for the hyper-parameters we use, that is $\lambda_{1}$ in Eq.~\ref{eq:st1}, $\alpha$, $\lambda_{2}$ and $K$ in Eq.~\ref{eq:st2}. As shown in Tab.~\ref{tab:ab_parameter}, $\lambda_{1} = 1e-4$, $\alpha = 10$, $\lambda_{2} = 1.0$ and $K = 2$ yields the best results on this dataset and we used the same values for all others.  Note that $\alpha = 10$ delivers much better performance than $\alpha=0$, which confirms that throwing away as few as 10\% of the pseudo labels makes a very significant difference. 

\parag{Stochastic Density Map.} 

To test if generating a stochastic density map instead of a deterministic one has a significant impact of performance, we compare the performance of \base{} that generates a deterministic map with a version of it that includes Masksembles to generate a stochastic map but still without any domain adaptation.  As can be seen in  Tab.~\ref{tab:ab_density}, the version with Masksembles does slightly better but not by a significant amount. Therefore, Masksembles by itself does not account for the large improvements we saw in Tab.~\ref{tab:eva}. 

\input{table/ab_density}

\parag{Choice of Auxiliary tasks.} 

Having chosen to use inverted images to provide a self-supervision signal may seem arbitrary during the first phase of training. To show that it is not, we tried variants in which we flip  the images left-right  (\ourm{}), we rotate them by $90$ degrees (\ourn{}) and by $270$ degrees (\ourt{}). As can be seen in Tab.~\ref{tab:ab_auxiliary}, \ourm{} performs on par with \ourp{}, the model trained without any image-wise supervision. \ourn{} and \ourt{} do slightly better but \ours{} is clearly best. This confirms the importance of flipping the images upside-down, which helps the network deal with perspective effects.

\input{table/ab_auxiliary}

\parag{Uncertainty Estimation.}

\input{table/ab_uncertain}

We use Masksembles~\cite{Durasov21} for uncertainty estimation because of its effectiveness and simplicity. However we could also have used MC-Dropout~\cite{Gal16} or Deep Ensembles~\cite{Lakshminarayanan17}. We tested both and report the results in Tab.~\ref{tab:ab_uncertain}. In addition to the usual MAE and RMSE, we also computed the {\it Pearson Correlation Coefficient}
\begin{align}
    r_{au}  & =  \frac{\sum_{i=1}^{n}(a_{i}-\bar{a})(u_{i}-\bar{u})}{ \sqrt{\sum_{i=1}^{n}(a_{i}-\bar{a})^2} \sqrt{\sum_{i=1}^{n}(u_{i}-\bar{u})^2}} \; ,     \label{eq:corr} \\
    \bar{a} & = \frac{1}{n}\sum_{i=1}^{n}a_{i} \; \mbox{and} \; \bar{u} = \frac{1}{n}\sum_{i=1}^{n}u_{i}   \; ,   \nonumber
\end{align}
where $n$ is the sample size, $a_{i},u_{i}$ are pixel-wise samples of counting error and uncertainty value respectively. $r_{au} \in [-1,1]$ and the higher its value is, the more correlated uncertainty is to the MAE error. In other words, when $r_{au}$ is large, it makes sense to discard uncertain densities as probably wrong and not to be used as pseudo labels. As can be seen in Tab.~\ref{tab:ab_uncertain}, using Masksembles~\cite{Durasov21}  as in \ours{} clearly outperform MC-Dropout~\cite{Gal16} and is comparable with Deep Ensembles~\cite{Lakshminarayanan17}. However, training Ensembles takes three times longer, which motivates our use of Masksembles.

%% file: fig/vis.tex

\begin{figure*}[t]
\centering
\begin{tabular}{ccc}
\includegraphics[width=.3\linewidth]{images/vis/img.jpg}&
\includegraphics[width=.3\linewidth]{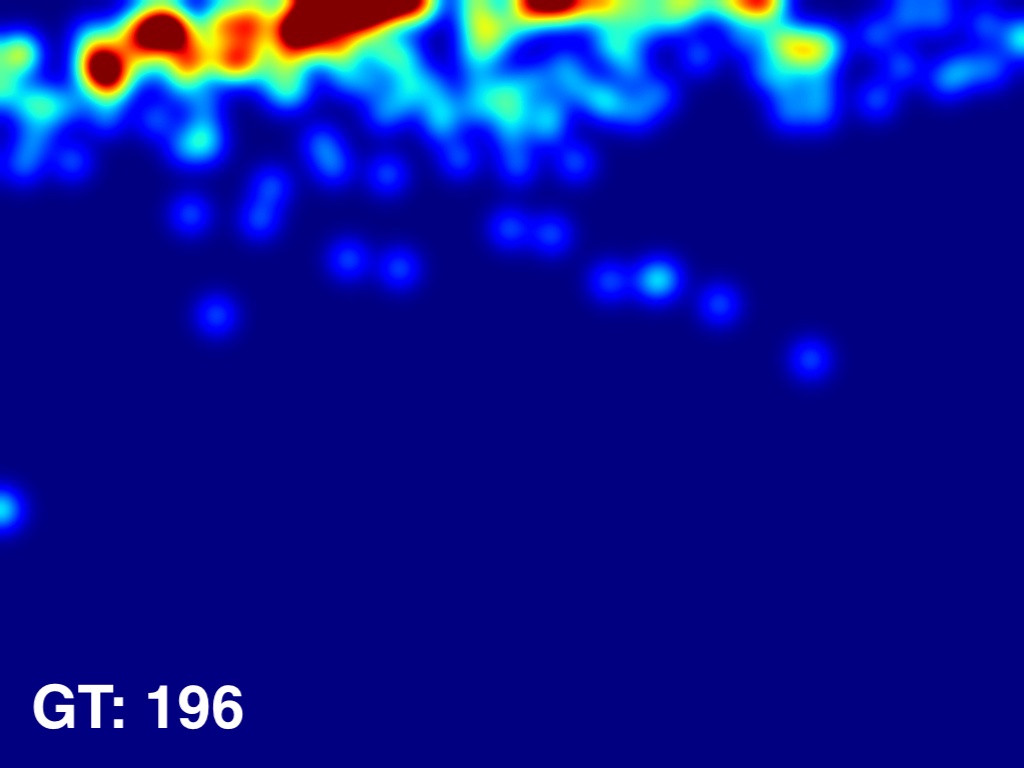}&
\includegraphics[width=.3\linewidth]{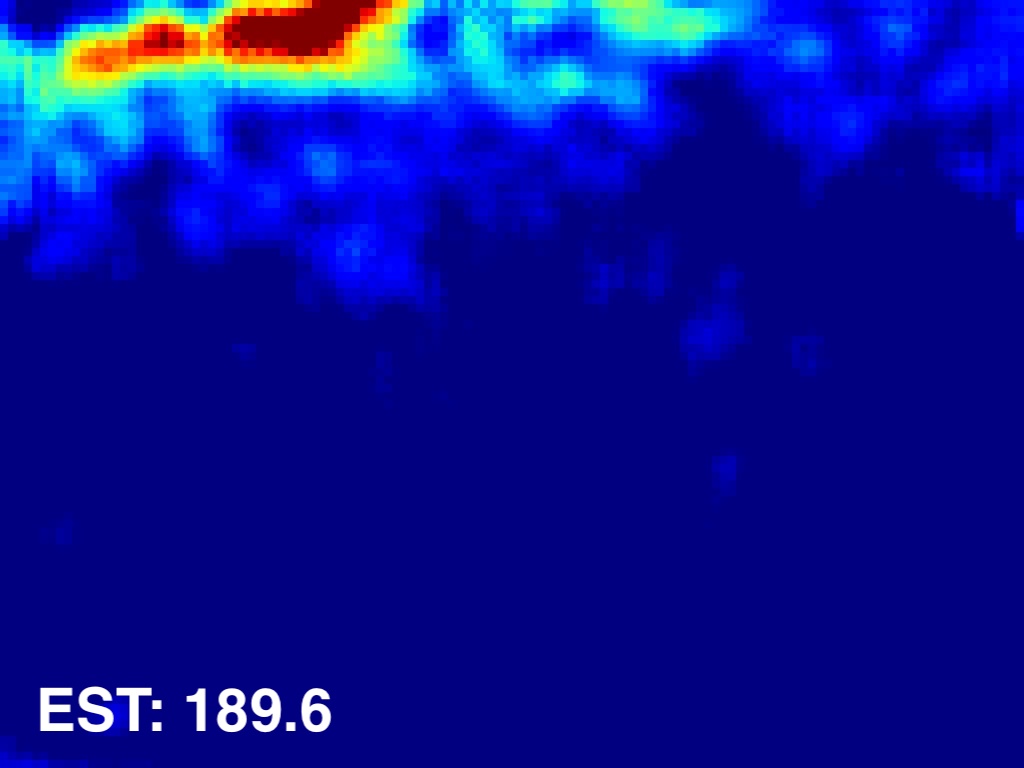}\\
\includegraphics[width=.3\linewidth]{images/vis/img_2.jpg}&
\includegraphics[width=.3\linewidth]{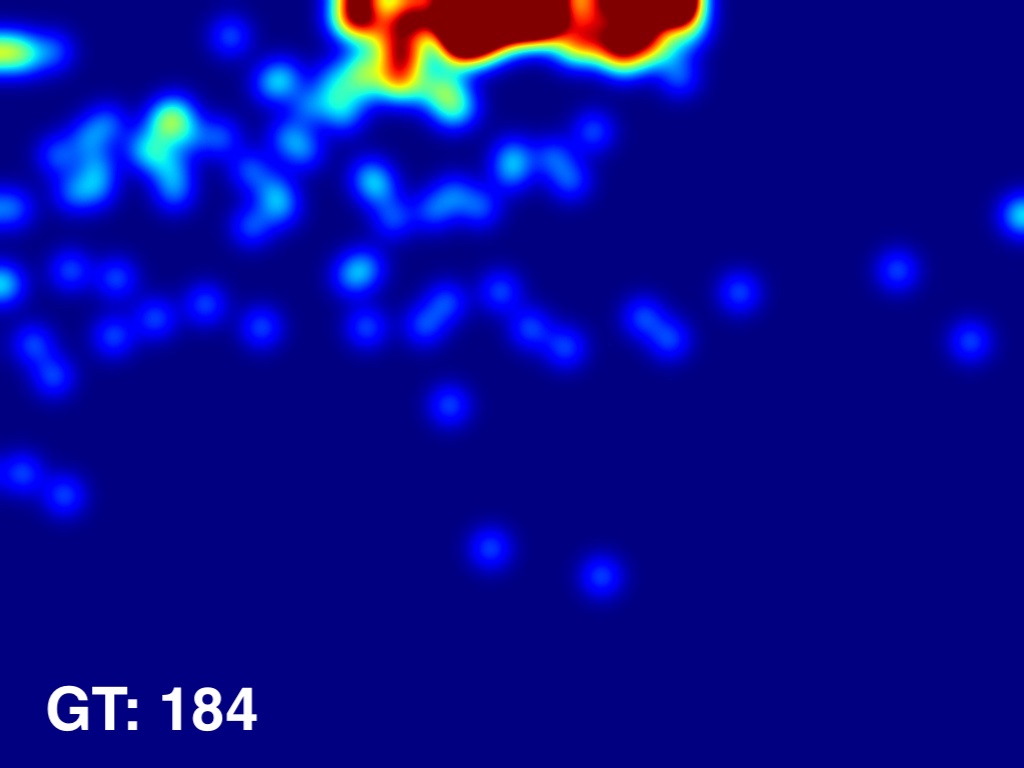}&
\includegraphics[width=.3\linewidth]{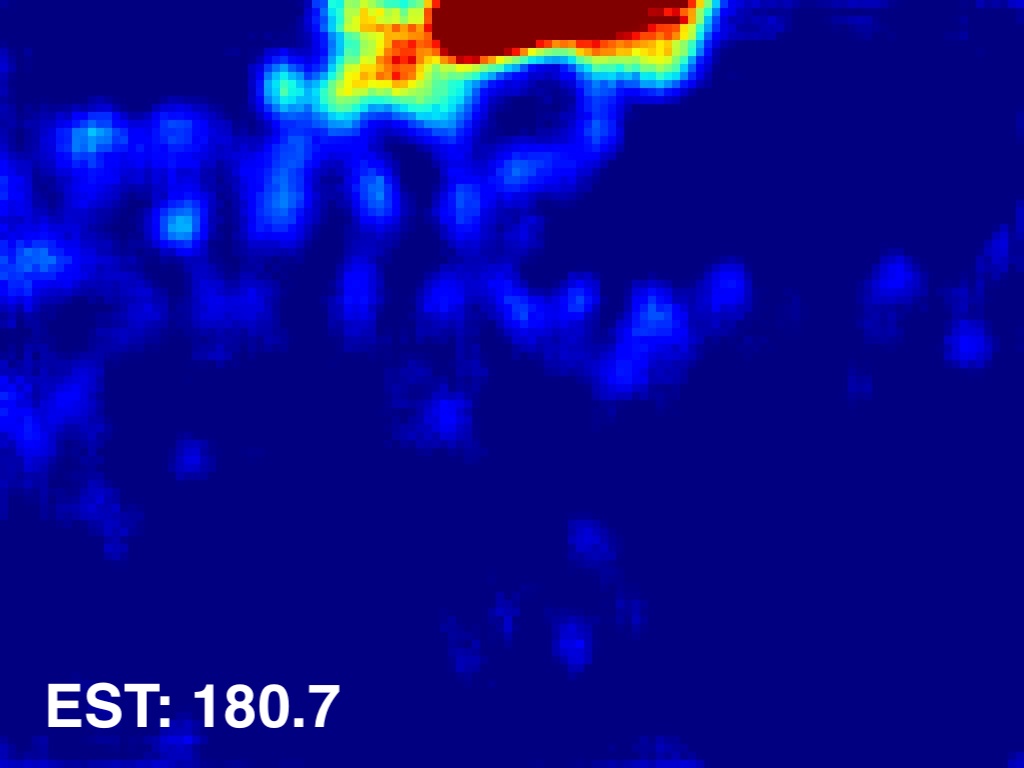}\\
\footnotesize{Input Image}&
\footnotesize{Ground Truth}&
\footnotesize{Estimated People Density}
\end{tabular}
  \caption{ {\bf Density maps.} We indicate the ground-truth and estimated total number of people in the bottom left corner of the density maps. Note how close our estimations are to the ground truth ones. Please refer to the supplementary material for additional such images.
 }
  \label{fig:vis}
  \end{figure*}

%% file: table/eva.tex

\begin{table*}[t]
  \begin{tabular}{ccc}
    \begin{minipage}{.3\linewidth}
  \centering
  \scalebox{0.8}{
    \rowcolors{2}{white}{gray!10}
    \begin{tabular}{lcc}
      \toprule
  Model  & $MAE$ & $RMSE$  \\
  \midrule
  No Adapt& 160.0 & 216.5  \\
  Cycle-GAN~\cite{Zhu17a} & 143.3 & 204.3 \\
  SE Cycle-GAN~\cite{Wang19a} & 123.4 & 193.4\\
  SE Cycle-GAN(JT)~\cite{Wang19d} & 119.6 & 189.1\\
  SE+FD~\cite{Han20} & 129.3 & 187.6\\
  GP~\cite{Sindagi20a}& 121 & 181 \\
  \ours{} & {\bf 109.2} & {\bf 168.1}  \\
  \bottomrule
  \end{tabular}
  }
\end{minipage} &

\begin{minipage}{.3\linewidth}
  \centering
  \scalebox{0.8}{
    \rowcolors{2}{white}{gray!10}
    \begin{tabular}{lcc}
      \toprule
      Model  & $MAE$ & $RMSE$  \\
      \midrule
      No Adapt& 22.8 & 30.6  \\
      Cycle-GAN~\cite{Zhu17a} & 25.4 & 39.7\\
      SE Cycle-GAN~\cite{Wang19a} & 19.9 & 28.3\\
      SE Cycle-GAN(JT)~\cite{Wang19d} & 16.4 & 25.8\\
      SE+FD~\cite{Han20} & 16.9 & 24.7\\
      GP~\cite{Sindagi20a}& 12.8 & 19.2\\
      \ours{} & {\bf 11.4} & {\bf 17.3}  \\
      \bottomrule
      \end{tabular}
      }
\end{minipage} &

\begin{minipage}{.3\linewidth}
  \centering
  \scalebox{0.8}{
    \rowcolors{2}{white}{gray!10}
    \begin{tabular}{lcc}
      \toprule
      Model  & $MAE$ & $RMSE$  \\
      \midrule
      No Adapt&  487.2 & 689.0  \\
      Cycle-GAN~\cite{Zhu17a} & 404.6 & 548.2 \\
      SE Cycle-GAN~\cite{Wang19a} & 373.4 & 528.8 \\
      SE Cycle-GAN(JT)~\cite{Wang19d} & 370.2 & 512.0\\
      GP~\cite{Sindagi20a}&  355 & 505 \\
      \ours{} & {\bf 336.5} & {\bf 486.1}  \\
      \bottomrule
      \end{tabular}
      }
\end{minipage} \\

\footnotesize{(a)}&
\footnotesize{(b)} &
\footnotesize{(c)} \\
\begin{minipage}{.3\linewidth}
  \centering
  \scalebox{0.8}{
    \rowcolors{2}{white}{gray!10}
    \begin{tabular}{lcc}
      \toprule
  Model  & $MAE$ & $RMSE$  \\
  \midrule
  No Adapt& 275.5 &  458.5 \\
  Cycle-GAN~\cite{Zhu17a} & 257.3 & 400.6\\
  SE Cycle-GAN~\cite{Wang19a} & 230.4 & 384.5\\
  SE Cycle-GAN(JT)~\cite{Wang19d} & 225.9 & 385.7\\
  SE+FD~\cite{Han20} & 221.2 & 390.2\\
  GP~\cite{Sindagi20a}& 210 & 351\\
  \ours{} & {\bf 198.3} & {\bf 332.9}  \\
  \bottomrule
  \end{tabular}
  }
\end{minipage} &

\multicolumn{2}{c}{
  \begin{minipage}{.5\linewidth}
    \centering
    \scalebox{0.8}{
      \rowcolors{2}{white}{gray!10}
      \begin{tabular}{lccccc|c}
        \toprule
      Model  & Scene1 & Scene2 & Scene3 & Scene4 & Scene5 &{\bf Average} \\
      \midrule
      No Adapt & 4.4 & 87.2 &  59. 1 & 51.8 & 11.7 & 42.8 \\
      Cycle-GAN~\cite{Zhu17a} & 4.4 & 69.6 & 49.9 & 29.2 & 9.0 & 32.4\\
      SE Cycle-GAN~\cite{Wang19a} & 4.3 & 59.1 & 43.7 & 17.0 &7.6 & 26.3\\
      SE Cycle-GAN(JT)~\cite{Wang19d} & 4.2 & 49.6 & 41.3 & 19.8 &7.2 & 24.4\\
      GP~\cite{Sindagi20a}& - & - & - & - & - & 20.4 \\
      \ours{} & {\bf 4.0} & {\bf 31.9} & {\bf 23.5} & {\bf 19.4} & {\bf 4.2} & {\bf 16.6}\\
       \bottomrule
      \end{tabular}
    }
  \end{minipage} 

} \\

\footnotesize{(d)}&
\multicolumn{2}{c}{\footnotesize{(e)}}
\end{tabular}
\vspace{2mm}
\caption{ {\bf Comparative results on different datasets.}  (a) {\bf ShanghaiTech Part A}. (b) {\bf ShanghaiTech Part B}. (c) {\bf UCF\_CC\_50}. (d) {\bf UCF-QNRF}. (e) {\bf WorldExpo'10}. Our approach consistently and clearly outperforms previous state-of-the-art methods on all the datasets.}
\label{tab:eva}
\end{table*}

%% file: table/ab_sup.tex

\begin{table}[t]
  \centering
  \scalebox{0.65}{
    \rowcolors{2}{white}{gray!10}
    \begin{tabular}{ccccccc}
      \toprule
        & \multicolumn{4}{c}{Self-Supervision}  &  &   \\
  Model  & Image & Synthetic Image& Pixel & 2nd Image& $MAE$ & $RMSE$  \\
  \midrule
  \base{}& & & & & 275.5 & 458.5  \\
  \ouri{} &\checkmark & & & & 242.8 & 407.6 \\
  \ouris{} &\checkmark &\checkmark & & & 243.0 & 406.8 \\
  \ourp{} & & & \checkmark& & 208.3 & 346.9 \\
  \ours{} &\checkmark & & \checkmark& & {\bf 198.3} & 332.9  \\
  \ourd{}& \checkmark & & \checkmark & \checkmark& 198.5 & {\bf 331.7} \\
  \bottomrule
    \end{tabular}
  }
\vspace{1mm}
\caption{\small {\bf Ablation study on self-supervision.} Both image-wise and pixel-wise self-supervision boost the performance and combining both further improves performance. By contrast, using image-wise self-supervision during the second stage, as opposed to the first, makes no obvious difference. }
\label{tab:ab_supervision}
\end{table}

%% file: table/ab_parameter.tex

\begin{table}[t]
  \begin{tabular}{cc}
    \begin{minipage}{.45\linewidth}
  \centering
  \scalebox{0.8}{
    \rowcolors{2}{white}{gray!10}
    \begin{tabular}{ccc}
      \toprule
  $\lambda_{1}$  & $MAE$ & $RMSE$ \\
  \midrule
   $1e-3$ & 208.0 & 344.2 \\
   $1e-4$ & {\bf 198.3} & {\bf 332.9}  \\
   $1e-5$ & 205.4 & 340.6  \\
  \bottomrule
    \end{tabular}
  }
\end{minipage} &
\begin{minipage}{.45\linewidth}
  \centering
  \scalebox{0.8}{
    \rowcolors{2}{white}{gray!10}
    \begin{tabular}{ccc}
      \toprule
  $\alpha$  & $MAE$ & $RMSE$ \\
  \midrule
   $0$ & 229.3 & 395.1 \\
   $5$ & 213.7 & 372.2 \\
   $10$ & {\bf 198.3} & {\bf 332.9}  \\
   $15$ &220.5 & 386.1  \\
  \bottomrule
    \end{tabular}
  }
\end{minipage}\\
\footnotesize{(a)}&
\footnotesize{(b)} \\

\begin{minipage}{.45\linewidth}
  \centering
  \scalebox{0.8}{
    \rowcolors{2}{white}{gray!10}
    \begin{tabular}{ccc}
      \toprule
  $\lambda_{2}$  & $MAE$ & $RMSE$ \\
  \midrule
   $0.5$ & 206.5 & 347.2 \\
   $1$ & {\bf 198.3} & {\bf 332.9}  \\
   $1.5$ & 204.9 & 350.7 \\
  \bottomrule
    \end{tabular}
  }
\end{minipage} &
\begin{minipage}{.45\linewidth}
  \centering
  \scalebox{0.8}{
    \rowcolors{2}{white}{gray!10}
    \begin{tabular}{ccc}
      \toprule
  $K$  & $MAE$ & $RMSE$ \\
  \midrule
   $1$ & 204.8 & 344.1 \\
   $2$ & {\bf 198.3} &  332.9 \\
   $3$ & 199.4 & {\bf 331.4}  \\
  \bottomrule
    \end{tabular}
  }
\end{minipage}\\
\footnotesize{(c)}&
\footnotesize{(d)}
\end{tabular}

\vspace{2mm}
\caption{ {\bf Ablation study on hyper-parameters.} $\lambda_{1} = 1e-4$, $\alpha = 10$, $\lambda_{2} = 1.0$ and $K = 2$ achieves the best performance, we thus use this setting for all the experiments. }
\label{tab:ab_parameter}
\end{table}

%% file: table/ab_density.tex

\begin{table}[t]
  \centering
  \scalebox{0.8}{
    \rowcolors{2}{white}{gray!10}
    \begin{tabular}{ccc}
      \toprule
  Model  & $MAE$ & $RMSE$  \\
  \midrule
  \base{} & 275.5 & 458.5  \\
  \base{}+Masksembles& {\bf 273.1} & {\bf 447.9} \\
  \bottomrule
    \end{tabular}
  }
\vspace{2mm}
\caption{ {\bf Ablation study on stochastic density map.} Generating stochastic density map slightly improve the performance but not by a significant amount. }
\label{tab:ab_density}
\end{table}

%% file: table/ab_auxiliary.tex

\begin{table}[t]
  \centering
  \scalebox{0.8}{
    \rowcolors{2}{white}{gray!10}
    \begin{tabular}{ccc}
      \toprule
  Model  & $MAE$ & $RMSE$  \\
  \midrule
  \ourp{}  & 208.3 & 346.9  \\
  \ourm{} & 208.1 & 346.0  \\
  \ourn{} & 205.5 & 344.7  \\
  \ourt{} & 204.8 & 342.1  \\
  \ours{} & {\bf 198.3} & {\bf 332.9}  \\
  \bottomrule
    \end{tabular}
  }
\vspace{2mm}
\caption{ \small {\bf Ablation study on auxiliary task.}  We tested different auxiliary tasks for image-wise supervision.  Flipping the image upside-down yields the best performance and we used it for all other experiments.}
\label{tab:ab_auxiliary}
\end{table}

%% file: table/ab_uncertain.tex

\begin{table}[t]
  \centering
  \scalebox{0.8}{
    \rowcolors{2}{white}{gray!10}
    \begin{tabular}{ccccc}
      \toprule
  Model  & Extra Cost& Correlation & $MAE$ & $RMSE$  \\
  \midrule
  MC-Dropout~\cite{Gal16}& & 0.18 & 209.8 & 344.9  \\
  Deep Ensembles~\cite{Lakshminarayanan17} &\checkmark & 0.44& 199.7& {\bf 331.8}   \\
  \ours{} & & {\bf 0.46} &  {\bf 198.3} & 332.9  \\
  \bottomrule
\end{tabular}
  }
\vspace{1mm}
\caption{ {\bf Ablation study on uncertainty estimation.} The Masksembles approach we used in measuring model uncertainty achieves better performance than MC-Dropout and similar performance as Deep Ensembles in terms of all three measures, and at a much lower computational cost.  }
\label{tab:ab_uncertain}
\end{table}

%% file: tex/conclusion.tex

\section{Conclusion}

We have proposed an approach to combining image-wise and pixel-wise self-supervision to substantially increase cross-domain crowd counting performance when only annotations of synthetic image is available. However, our approach does not require the source images to be synthetic and could take advantage of additional annotations when available. In future work, we will therefore expand it to using multiple datasets of real-world images with partial annotations.

\vspace{-0.05em}
{\bf Acknowledgments} This work was supported in part by the Swiss National Science Foundation.